\documentclass[conference]{IEEEtran}
\IEEEoverridecommandlockouts
\usepackage{cite}
\usepackage{amsmath,amssymb,amsfonts}
\usepackage{algorithm}
\usepackage{algorithmic}
\usepackage{geometry}
\usepackage{graphicx}
\usepackage{textcomp}
\usepackage{xcolor}
\usepackage{hyperref}
\hypersetup{
    colorlinks=true,
    citecolor=black,
    linkcolor=black,
    filecolor=black,      
    urlcolor=black,
    pdftitle={Online outlier Detection Threshold},
    pdfauthor={Marek Wadinger}
    pdfpagemode=FullScreen,
    }
\def\BibTeX{{\rm B\kern-.05em{\sc i\kern-.025em b}\kern-.08em
    T\kern-.1667em\lower.7ex\hbox{E}\kern-.125emX}}
\begin{document}

\title{\vspace{0.25in}Adaptable and Interpretable Framework for Novelty Detection in Real-Time IoT Systems\\
\thanks{The Authors gratefully acknowledge the contribution of the Slovak Research and Development Agency under the project APVV-20-0261. The authors gratefully acknowledge the contribution of the Scientific Grant Agency of the Slovak Republic under the grant 1/0490/23. This research is funded by the European Commission under the grant no. 101079342 (Fostering Opportunities Towards Slovak Excellence in Advanced Control for Smart Industries).}
}

\author{\IEEEauthorblockN{Marek Wadinger and Michal Kvasnica}
\IEEEauthorblockA{\textit{Institute of Information Engineering, Automation and Mathematics} \\
\textit{Slovak University of Technology in Bratislava}\\
Bratislava, Slovakia \\
\{marek.wadinger, michal.kvasnica\}@stuba.sk}
}

\maketitle

\begin{abstract}
This paper presents the Real-time Adaptive and Interpretable Detection 
This paper presents the Real-time Adaptive and Interpretable Detection (RAID) algorithm. The novel approach addresses the limitations of state-of-the-art anomaly detection methods for multivariate dynamic processes, which are restricted to detecting anomalies within the scope of the model training conditions. The RAID algorithm adapts to non-stationary effects such as data drift and change points that may not be accounted for during model development, resulting in prolonged service life. A dynamic model based on joint probability distribution handles anomalous behavior detection in a system and the root cause isolation based on adaptive process limits. RAID algorithm does not require changes to existing process automation infrastructures, making it highly deployable across different domains. Two case studies involving real dynamic system data demonstrate the benefits of the RAID algorithm, including change point adaptation, root cause isolation, and improved detection accuracy.
\end{abstract}

\begin{IEEEkeywords}
Fault diagnosis; Iterative learning control; Statistical learning
\end{IEEEkeywords}

\section{Introduction}\label{Introduction}
Anomaly detection systems, able to discriminate abnormal, unexpected patterns and adapt to novel expected patterns in data, are known to be an essential part of risk-averse systems. In particular, anomaly detection systems assess the normal operational conditions allowing Internet of Things (IoT) devices to stream high-fidelity data into control units.

In their highly influential paper, Chandola et al. review former research efforts spanning diverse application domains \cite{Chandola2009}.
Recent studies highlight the need to develop holistic methods with general application and accessible tunability for operators \cite{Laptev2015}, \cite{Kejariwal2015}, \cite{Cook2020}. 

Cook et al. denote substantial aspects that pose challenges to anomaly detection on IoT, namely the context information of the measurement being temporal, spatial, and external, multivariate character, noise, and nonstationarity \cite{Cook2020}. Feature engineering methods allow encoding contextual properties and increase the performance \cite{Fan2019}. However, extensive feature engineering may significantly increase dimensionality, requiring sizeable data storage and high computational resources \cite{Talagala2021}. 

Moreover, nonstationarity resulting from concept drift, an alternation in the pattern of data due to a change in statistical distribution, and change points, permanent changes to the system's state, represents a difficulty of a significant extent. In real-world scenarios, those changes are frequently unpredictable. Therefore, the ability of an anomaly detection method to adapt to changes in the data structure is crucial for long-term deployments. The former scalability problem now introduces a significant latency in detector adaptation \cite{Wu2021}. Incremental learning methods allowed adaptation while restraining the storage of the whole dataset. The supervised operator-in-the-loop solution offered by Pannu et al. showed the detector's adaptation to data labeled on the flight. 
Others approached the problem as sequential processing of bounded data buffers in univariate signals \cite{Ahmad2017134} and multivariate systems \cite{Bosman201514}. 

Lastly, recent efforts to extend anomaly detection tasks to root cause isolation governed the development of explanatory methods capable of diagnosing and tracking faults across the system. Studies can be split into two groups. The first group approaches explainability as the importance of individual features \cite{Carletti2019}, \cite{Nguyen2019}, \cite{Amarasinghe2018}. Those studies allow an explanation of novelty by considering features independently. The second group uses statistical learning creating models explainable via probability. Yang et al. recently proposed a Bayesian network (BN) for fault detection and diagnosis task. Individual nodes of the network represent normally distributed variables, whereas the multiple regression model defines weights and relationships. Using the predefined structure of the BN, the authors propose an offline-trained model with online detection and diagnosis \cite{Yang2022}. Offline training, however, as we wrote earlier, do not allow adaptation to expected novel pattern and, therefore, to our knowledge, is not suitable for long-term operation on real IoT devices.

This paper emphasizes the importance of such adaptability in anomaly detection and proposes a method that addresses this challenge. Here we report the discovery and characterization of an adaptive anomaly detection method for streaming IoT data. The ability to diagnose multivariate data while providing root cause isolation, inherent in the univariate case, extends our previous contribution to the field as presented in \cite{Wadinger2023}. The proposed algorithm represents a general method for a broad range of safety-critical systems where anomaly diagnosis and identification is crucial.

Two case studies show that the proposed method based on dynamic joint normal distribution gives the capacity to explain novelties and isolate the root cause of anomalies and allow adaptation to change points advancing recently developed anomaly detection techniques to the long-term deployment of the service and cross-domain usage. We observe similar detection performance for the cost of lower scalability.

The main contribution of the proposed solution to the developed body of research is that it:
\begin{itemize}
\item Provides both adaptability and interpretability
\item Identifies systematic outliers and root cause
\item Uses self-learning approach on streamed data
\item Utilizes existing IT infrastructure
\item Establishes dynamic limits for signals
\end{itemize}

\section{Preliminaries}
In this section, we present the fundamental ideas that form the basis of the developed approach. Subsection \ref{AA:Welford} explains Welford's online algorithm, which can adjust distribution to changes in real time. Subsection \ref{AA:InvWelford} proposes a two-pass implementation that can reverse the impact of expired samples. The math behind distribution modeling in Subsection \ref{AA:Distribution} establishes the foundation for the Gaussian anomaly detection model discussed in the final Subsection \ref{AA:Anomaly} of the preliminaries.

\subsection{Welford's Online Algorithm}\label{AA:Welford}
Welford introduced a numerically stable online algorithm for calculating mean and variance in a single pass. The algorithm allows the processing of IoT device measurements without the need to store their values \cite{Wel62}.

Given measurement \(x_i\) where \(i=1,...,n\) is a sample index in sample population \(n\), the corrected sum of squares \(S_n\) is defined as
\begin{equation}
S_n = \sum_{i=1}^n (x_i - \bar x_n)^2\text{,}\label{eq:sumsquares}
\end{equation}
with the running mean \(\bar x_n\) defined as previous mean \(\bar x_{n-1}\) weighted by proportion of previously seen population \(n-1\) corrected by current sample as
\begin{equation}
\bar x_n = \frac{n-1}{n} \bar x_{n-1} + \frac{1}{n}x_n = \bar x_{n-1} + \frac{x_n - \bar x_{n-1}}{n}\text{.}\label{eq:runmean}
\end{equation}
Throughout this paper, we consider a following formulation of an update to the corrected sum of squares:
\begin{equation}
S_n = S_{n-1} + (x_n - \bar x_{n-1})(x_n - \bar x_n)\text{,}\label{eq:upsumsquares}
\end{equation}
as it is less prone to numerical instability due to catastrophic cancellation. Finally, the corresponding unbiased variance is
\begin{equation}
s^2_n = \frac{S_{n}}{n-1}\text{.}\label{eq:runvar}
\end{equation}

This implementation of the Welford method requires the storage of three scalars: \(\bar x_{n-1}\); \(n\); \(S_n\).

\subsection{Inverse Welford's Algorithm}\label{AA:InvWelford}
Based on \eqref{eq:runmean}, it is clear that the influence of the latest sample over the running mean decreases as the population \(n\) grows. For this reason, regulating the number of samples used for sample mean and variance computation has crucial importance over adaptation. Given access to the instances used for computation and expiration period \(t_e \in \mathbb{N}_{0}^{n-1}\), reverting the impact of \(x_{n-t_e}\) can be written as follows

\begin{equation}
S_{n-1} = S_n - (x_{n-t_e} - \bar x_{n-1})(x_{n-t_e} - \bar x_n)\text{,}\label{eq:revrunmean}
\end{equation}

where the reverted mean is given as

\begin{equation}
\bar x_{n-1} = \frac{n}{n-1} \bar x_{n} - \frac{1}{n-1}x_{n-t_e} = \bar x_{n} - \frac{x_{n-t_e} - \bar x_{n}}{n-1}\text{.}\label{eq:revmean}
\end{equation}

Finally, the unbiased variance follows the formula:

\begin{equation}
s^2_{n-1} = \frac{S_{n-1}}{n-2}\text{.}\label{eq:revvar}
\end{equation}

\subsection{Statistical Model of Multivariate System}\label{AA:Distribution}
Multivariate normal distribution generalizes the multivariate systems to the model where the degree to which variables are related is represented by the covariance matrix. Gaussian normal distribution of variables is a reasonable assumption for process measurements, as it is a common distribution that arises from stable physical processes measured with noise. The general notation of multivariate normal distribution is:
\begin{equation}\mathbf{X}\ \sim\ \mathcal{N}_k(\boldsymbol\mu,\, \boldsymbol\Sigma)\text{,}
\end{equation}

where $k$-dimensional mean vector is denoted as \(\boldsymbol\mu = (\bar x_{1},...,\bar x_{k})^T\ \in \mathbb{R}^{k}\) and \(\boldsymbol\Sigma \in \mathbb{R}^{k\times{k}}\) is the $k \times k$ covariance matrix, where \(k\) is the index of last random variable.

The probability density function (PDF) \(f(\boldsymbol{x}; \boldsymbol{\mu}, \boldsymbol{\Sigma})\) of multivariate normal distribution is denoted as:
\begin{equation}
f(\boldsymbol{x}; \boldsymbol{\mu}, \boldsymbol{\Sigma}) = \frac{1}{(2\pi)^{k/2} |\boldsymbol{\Sigma}|^{1/2}} e^{-\frac{1}{2} (\boldsymbol{x}-\boldsymbol{\mu})^\top \boldsymbol{\Sigma}^{-1} (\boldsymbol{x}-\boldsymbol{\mu})}\text{,}
\end{equation}

where $\boldsymbol{x}$ is a $k$-dimensional vector of measurements $x_i$ at time $i$, $|\boldsymbol{\Sigma}|$ denotes the determinant of $\boldsymbol{\Sigma}$, and $\boldsymbol{\Sigma}^{-1}$ is the inverse of $\boldsymbol{\Sigma}$.

The cumulative distribution function (CDF) of a multivariate Gaussian distribution describes the probability that all components of the random matrix \(\boldsymbol{X}\) take on a value less than or equal to a particular point \(\boldsymbol{x}\) in space, and can be used to evaluate the likelihood of observing a particular set of measurements or data points. The CDF is often used in statistical applications to calculate confidence intervals, perform hypothesis tests, and make predictions based on observed data. In other words, it gives the probability of observing a random vector that falls within a certain region of space. The standard notation of CDF is as follows:

\begin{equation}
F(\boldsymbol{x}; \boldsymbol{\mu}, \boldsymbol{\Sigma}) = \int_{-\infty}^{\boldsymbol{x}} f(\boldsymbol{x}; \boldsymbol{\mu}, \boldsymbol{\Sigma})  \text{d}\boldsymbol{x}\text{,}\label{eq:cdf}
\end{equation}

where $\text{d}\boldsymbol{x}$ denotes the integration over all $k$ dimensions of $\boldsymbol{x}$. 

As the equation \eqref{eq:cdf} cannot be integrated explicitly, an algorithm for numerical computation was proposed in \cite{Genz2000}.

Given the PDF, we can also determine the value of \(\boldsymbol{x}\) that corresponds to a given quantile $q$  using a numerical method for inversion of CDF (ICDF) often denoted as percent point function (PPF) or $F(\boldsymbol{x}; \boldsymbol{\mu}, \boldsymbol{\Sigma})^{-1}$. An algorithm that calculates the value of the PPF for univariate normal distribution is reported below as Algorithm \ref{alg:ppf}.

\begin{algorithm}[H]
\caption{{Percent-Point Function for Normal Distribution}} \label{alg:ppf}
 \begin{algorithmic}[1]
 \renewcommand{\algorithmicrequire}{\textbf{Input:}}
 \renewcommand{\algorithmicensure}{\textbf{Output:}}
 \REQUIRE quantile $q$, sample mean $\bar x_n$ \eqref{eq:runmean}, sample variance $s^2_n$ \eqref{eq:runvar}
 \ENSURE  threshold value $x_{n,q}$
 \\ \textit{Initialisation} : 
  \STATE $f \leftarrow 10$; $l \leftarrow -f $; $r \leftarrow f;$
 \\ \textit{LOOP Process}
  \WHILE {$F(l; \bar x_n, s^2_n) > 0$}
  \STATE $r \leftarrow l $;
  \STATE $l \leftarrow lf $;
  \ENDWHILE
  \WHILE {$F_X(r)-q < 0$}
    \STATE $l \leftarrow r $;
    \STATE $r \leftarrow rf $;
  \ENDWHILE
  \STATE {$\tilde{x}_{n,q} = \text{arg} \min_{x_n} \| F(x_n; \bar x_n, s^2_n) - q \| ~ \text{s.t.} ~ l \le x_n \le r$}
 \RETURN $\tilde{x}_{n,q}  \sqrt{s^2_n} + \bar x_n $
 \end{algorithmic}
\end{algorithm}

The Algorithm \ref{alg:ppf} for PPF computation is solved using an iterative root-finding algorithm such as Brent's method \cite{Brent72}.

\subsection{Multivariate Gaussian Anomaly Detection}\label{AA:Anomaly}
From a statistical viewpoint, outliers can be denoted 
as values that significantly deviate from the mean. Assuming that the spatial and temporal characteristics of the system over the moving window can be encoded as normally distributed features, we can claim, that any anomaly may be detected as an outlier.

In empirical fields, such as machine learning, the three-sigma rule ($3\sigma$) defines a region of distribution where normal values are expected to occur with near certainty. This assumption makes approximately 0.27\% of values in the given distribution considered anomalous. 

The \(3\sigma\) rule establishes the probability that any sample \(\boldsymbol{x_i}\) of a random variable \(\boldsymbol{X}\) lies within a given CDF over a semi-closed interval as the distance from the sample mean \(\boldsymbol{\mu}\) of 3 sample standard deviations \(\boldsymbol{\Sigma}\) and gives an approximate value of $q$ as
\begin{equation}
q=P\{|\boldsymbol{x_i}-\boldsymbol{\mu}|<3\boldsymbol{\Sigma}\}=0.99730\text{.}
\end{equation}

Using a probabilistic model of normal behavior lets us query the threshold vectors \(\boldsymbol{x_{l}}\) and \(\boldsymbol{x_{u}}\) which corresponds to the closed interval of CDF at which probability was established. Inversion of \eqref{eq:cdf} can be used for such query resulting in:

\begin{equation}
\boldsymbol{x_l} = F((1 - P\{|\boldsymbol{x_i}-\boldsymbol{\mu}|<3\boldsymbol{\Sigma}\circ\boldsymbol{I}\}); \boldsymbol{\mu}, \boldsymbol{\Sigma}\circ\boldsymbol{I})^{-1}\text{,}\label{eq:thresh_low}
\end{equation}

for the lower limit, and

\begin{equation}
\boldsymbol{x_u} = F((P\{|\boldsymbol{x_i}-\boldsymbol{\mu}|<3\boldsymbol{\Sigma}\circ\boldsymbol{I}\}); \boldsymbol{\mu}, \boldsymbol{\Sigma}\circ\boldsymbol{I})^{-1}\text{,}\label{eq:thresh_high}
\end{equation}

for upper one, where $\boldsymbol{\Sigma}\circ\boldsymbol{I}$ represents diagonal elements of $\boldsymbol{\Sigma}$.

However, the problem of computing CDF of a multivariate normal distribution is that it may result in numerical issues for small probabilities. To avoid underflow, the logarithm of CDF (log-CDF) is computed, converting the product of individual elements into a numerically more stable summation. The value of $T$ represents a threshold, defining the discrimination boundary between normal operation and anomaly. The predicted state of the system $Y_i$ at time $i$ is defined as
\begin{equation}
y_i =
  \begin{cases}
     0 & \text{ if } T \leq \log{F(\boldsymbol{x_i}; \boldsymbol{\mu}, \boldsymbol{\Sigma})} \\ 
     1 & \text{ if }  T > \log{F(\boldsymbol{x_i}; \boldsymbol{\mu}, \boldsymbol{\Sigma})}\text{,}\label{eq:anomaly}
  \end{cases}
\end{equation}

where $y_i = 0$ for normal operation of the system and $y_i = 1$ for anomalous operation.

\section{Novelty Detection and Interpretation Framework}
In this section, we propose a real-time adaptive and interpretable detection (RAID) scheme for multivariate systems with streaming IoT devices. The proposed approach models the system as dynamic joint normal distribution to allow adaptability to omnipresent nonstationary effects on processes, handling change points, concept drift, and seasonal effects. Our main contribution represents combining an adaptable self-supervised system with the root cause identification capability, providing the online statistical model with the capacity to diagnose anomalies in two ways. Firstly, by computing global log-probability to detect the system's operating conditions. Secondly, by isolating outliers in individual signal measurements and features based on dynamic alert-triggering process limits. In what follows, the method is divided into three parts and described in subsections. The first subsection focuses on the initialization of the model's parameters. The second subsection describes the process of online training and adaptation. The last subsection describes the prediction and diagnostic capabilities of the model. The Algorithm \ref{alg:detector} provides a simplified representation of the method.

\subsection{Model Parameters Initialization}\label{init}
The model initialization is governed by specifying tunable hyperparameters of the model: expiration period $t_e$ and threshold $T$. The expiration period sets the window size of time-rolling computations to change the proportion of outliers in the given time frame and directly influences the relaxation (longer expiration period) or tightening (shorter expiration period) of the dynamic signal limits. A grace period, defaulting to $3/4 t_e$, represent time sufficient for model calibration, during which outliers are not detected to avoid false positive identification and speed up the process of self-supervised learning introduced later in Subsection \ref{train}. The length of the expiration period is inversely related to the model's change point adaptation capacity, making it more robust to sudden changes. The ability of the model to adapt to changes in the underlying data-generating process, such as shifts in the mean or variance of the process, is handled via adaptation period $t_a$. Longer $t_a$ means slower adaptation for the cost of longer alerts which might be, however, valuable at times when unexpected outlier spans longer time frames. For most cases, the model works best for $t_a = 1/4 t_e$.

As a general rule of thumb, expiration period $t_e$ shall be chosen based on the slowest dynamics of the multivariate system. The existence of two tunable and easy-to-interpret hyper-parameters makes it very easy to adapt the solution to any multivariate system.

\subsection{Online training}\label{train}
Training on the data stream requires an incremental learning scheme, i.e., one sample at a time at the moment of its arrival. Incremental learning allows online adaptation, which refers to an ability to update the model's parameters on new observations. Such ability comes at the cost of computational delay, which may pose challenges concerning the latency of the detector's response. 

In the case of a dynamic joint probability distribution, the parameters are $\boldsymbol{\mu_i}$ and $\boldsymbol{\Sigma_i}$ at time instance \(i\). Update of the mean vector $\boldsymbol{\mu_i}$ and covariance matrix $\boldsymbol{\Sigma_i}$ is governed by Welford's online algorithm using equation \eqref{eq:runmean} and \eqref{eq:runvar} respectively. Samples after the expiration period $t_e$ are forgotten in the second pass. The effect of expired samples is reverted using inverse Welford's algorithm for mean \eqref{eq:revmean} and variance \eqref{eq:revvar}, accessing the data in the buffer. The expired samples are dropped in this second pass. For details, refer to Subsection \ref{AA:InvWelford}.

It is important to note that adaptation is performed in a self-supervised fashion. Previous routine runs if the observation at time instance \(i\) is considered normal. Adaptation period $t_a$ allows the model to update the distribution on outliers as well. Given the predicted system anomaly state from \eqref{eq:anomaly} as $y_i$  over the window of past observations \(\boldsymbol{y_i}=\{y_{i-t_a},...,y_{i}\}\), the following test holds when adaptation is performed on outlier:

\begin{equation}
{\frac{\sum_{y\in \boldsymbol{y_i}}y}{n(\boldsymbol{y_i})}} > 2*(q-0.5)\text{.}\label{eq:condition}
\end{equation}
where \(n(\boldsymbol{y_i})\) represents dimensionality of \(\boldsymbol{y_i}\). The logic of the \eqref{eq:condition} follows the probabilistic approach to anomalies that assumes a number of anomalies are lower or equal to the conditional probability at both tails of the distribution

\subsection{Online prediction}\label{predict}
In the prediction phase, multiple metrics are evaluated to assess the state of the modeled system. 

Global log-CDF of multivariate Gaussian distribution computed, using the numerical algorithm proposed in \cite{Genz2000}, for process observation vector $\boldsymbol{x_i}$ at time instance $i$, serves for the establishment of anomalous/normal behavior of the system as whole. The interactions between input signals and features are inherently considered. 

For root cause isolation, inputs are tested against the interval given by lower and upper threshold values, $\boldsymbol{x_l}$ from \eqref{eq:thresh_low} and $\boldsymbol{x_u}$ from \eqref{eq:thresh_high} respectively. Algorithm \ref{alg:ppf} is used to compute PPF for input at every time instance using updated parameters of the model. Lower and upper thresholds alone can be interpreted as dynamic process limits, updating conservative process limits provided by the sensor documentation, anticipating aging as well as environmental conditions influencing its operation.

Signal loss, an unexpected novel behavior, is anticipated within the framework computing the CDF over the univariate normal distribution of sampling, the differences between subsequent timestamps. We assume, that in the long run, the distribution of sampling times is not subject to drift. Therefore, one pass update using \eqref{eq:runmean} and \eqref{eq:runvar} is employed. To foresee subtle changes in sampling, self-supervised learning uses anomalies weighted by deviation of $(1 - F(\boldsymbol{x_i}; \boldsymbol{\mu}, \boldsymbol{\Sigma}))$ for training.

Change points are isolated when adaptation test from \eqref{eq:condition} hold true, triggering the update of the model.

\begin{algorithm}[H]
\caption{{Online Detection and Identification Workflow using RAID method}} \label{alg:detector}
 \begin{algorithmic}[1]
  \renewcommand{\algorithmicrequire}{\textbf{Input:}}
  \renewcommand{\algorithmicensure}{\textbf{Output:}}
  \REQUIRE expiration period $t_e$
  \ENSURE  system anomaly $y^g_i$, signal anomalies $\boldsymbol{y^s_i}$, sampling anomaly $y^t_i$, change-point $y^c_i$, lower thresholds $\boldsymbol{x_{l,i}}$, upper thresholds $\boldsymbol{x_{u,i}}$, 
 \\ \textit{Initialisation} : 
  \STATE $i \leftarrow 1;~ n \leftarrow 1;~ q \leftarrow 0.9973;~ \boldsymbol{\mu}  \leftarrow \boldsymbol{x_0};~  \boldsymbol{\Sigma} \leftarrow \mathbf{1}_{k \times k}$;
  \STATE compute $F(\boldsymbol{x_0}; \boldsymbol{\mu}, \boldsymbol{\Sigma})$ using algorithm in \cite{Genz2000};
 \\ \textit{LOOP Process}
  \LOOP
    \STATE {$\boldsymbol{x_i} \leftarrow$ RECEIVE()};
    \STATE $y^g_i \leftarrow$ PREDICT($\boldsymbol{x_i}$) using \eqref{eq:anomaly};
    \STATE $\boldsymbol{x_{l,i}}\text{, }\boldsymbol{x_{u,i}} \leftarrow$ GET($q, \boldsymbol{\mu}, \boldsymbol{\Sigma}$) using Algorithm \ref{alg:ppf};
    \STATE $\boldsymbol{y^s_i} \leftarrow$ INRANGE($\boldsymbol{x_i}, [\boldsymbol{x_{l,i}}\text{, }\boldsymbol{x_{u,i}}]$);
    \IF {\eqref{eq:anomaly} \OR \eqref{eq:condition}}
     \STATE {$\boldsymbol{\mu}, \boldsymbol{\Sigma} \leftarrow$ UPDATE($\boldsymbol{x_i}, \boldsymbol{\mu}, \boldsymbol{\Sigma}, n$) using \eqref{eq:runmean}, \eqref{eq:runvar}};
     \IF {\eqref{eq:condition}}
      \STATE $y^c_i \leftarrow 1$;
     \ELSE
      \STATE $y^c_i \leftarrow 0$;
     \ENDIF
     \STATE $n \leftarrow n + 1$;
     \FOR {$\boldsymbol{x_{i-t_e}}$}
      \STATE {$\boldsymbol{\mu}, \boldsymbol{\Sigma} \leftarrow$ REVERT($\boldsymbol{x_{i-t_e}}, \boldsymbol{\mu}, \boldsymbol{\Sigma}, n$) using \eqref{eq:revmean}, \eqref{eq:revvar}};
      \STATE $n \leftarrow n - 1$;
     \ENDFOR
    \ENDIF
    \STATE $i \leftarrow i + 1$;
  \ENDLOOP
 \end{algorithmic}
\end{algorithm}

\section{Case Study}
This section provides two case studies that showcase the effectiveness and applicability of our proposed approach. In these studies, we investigate the properties and performance of the approach using streamed benchmark system data and signals from IoT devices in a microgrid system.  The successful deployment demonstrates that this approach is suitable for existing process automation infrastructure.

The case studies were realized using Python 3.10.1 on a machine employing an 8-core Apple M1 CPU and 8 GB RAM.

\subsection{Benchmark}\label{AA:Benchmark}
In this subsection, we compare the proposed method with standard adaptive unsupervised detection methods. Two of the most well-known methods, providing iterative learning capabilities over multivariate time-series data are One-Class Support Vector Machine (OC-SVM) and Half Spaced Trees (HS-Trees). Both methods represent the state of the art for cases of anomaly detection on dynamic system data. Comparison is conducted on real benchmarking data, annotated with labels of whether the observation was anomalous or normal. The dataset of Skoltech Anomaly Benchmark (SKAB) \cite{skab2020} is used for this purpose. It represents a combination of experiments with the behavior of rotor imbalance as a subject to various functions introduced to control action as well as slow and sudden changes in the amount of water in the circuit. The system is described by 8 features. The data were preprocessed according to best practices for the given method, namely: standard scaling for OC-SVM, normalization for HS-Trees, and no scaling for the proposed method. The optimal quantile threshold value for both reference methods is found using grid search. Results are provided within Table \ref{tab:perf_comp}, evaluating F1 score, Recall and Precision. Value of 100\% at each metric represents a perfect detection. The latency represents the average computation time per sample of the pipeline including training and data preprocessing. 

\begin{table}[htbp]
  \caption{Metrics evaluation on SKAB dataset}
  \begin{center}
  \label{tab:perf_comp}
  \begin{tabular}{|c|c|c|c|c|}
    \hline
    Metric & RAID & OC-SVM & HS-Trees \\
    \hline
    F1 [$\%$] & $\boldsymbol{48.70}$ & 44.42 & 34.10 \\
    \hline
    Recall [$\%$] & 49.90 & $\boldsymbol{56.67}$ & 32.57 \\
    \hline
    Precision [$\%$] & $\boldsymbol{47.56}$ & 36.52 & 35.77 \\
    \hline
    Avg. Latency [ms] & 1.55 & 0.44 & $\boldsymbol{0.21}$ \\
    \hline
  \end{tabular}
  \end{center}
\end{table}

The results in Table \ref{tab:perf_comp}suggest that our algorithm provides slightly better performance than reference methods. Based on the Scoreboard for various algorithms on SKAB's Kaggle page, our iterative approach performs comparably to the evaluated batch-trained model. Such a model has all the training data available before prediction unlike ours, evaluating the metrics iteratively on a streamed dataset.

\subsection{Battery Energy Storage System (BESS)}\label{AA:BESS}
In the second case study, we verify our proposed method on BESS. The BESS reports measurements of State of Charge (SoC), supply/draw energy set-points, and inner temperature, at the top, middle, and bottom of BESS. Tight battery cell temperature control is needed to optimize performance and maximize the battery's lifespan. Identifying anomalous events and removal of corrupted data might yield significant improvement on the process control level. 

The default sampling rate of the signal measurement is 1 minute. However, network communication of the IoT devices is prone to packet dropout, which results in non-uniform sampling. The data are normalized to the range $[0, 1$] to protect the sensitive business value. The proposed approach is deployed to the existing infrastructure of the system, allowing real-time detection and diagnosis of the system.

\begin{figure}[htbp]
\centerline{\includegraphics{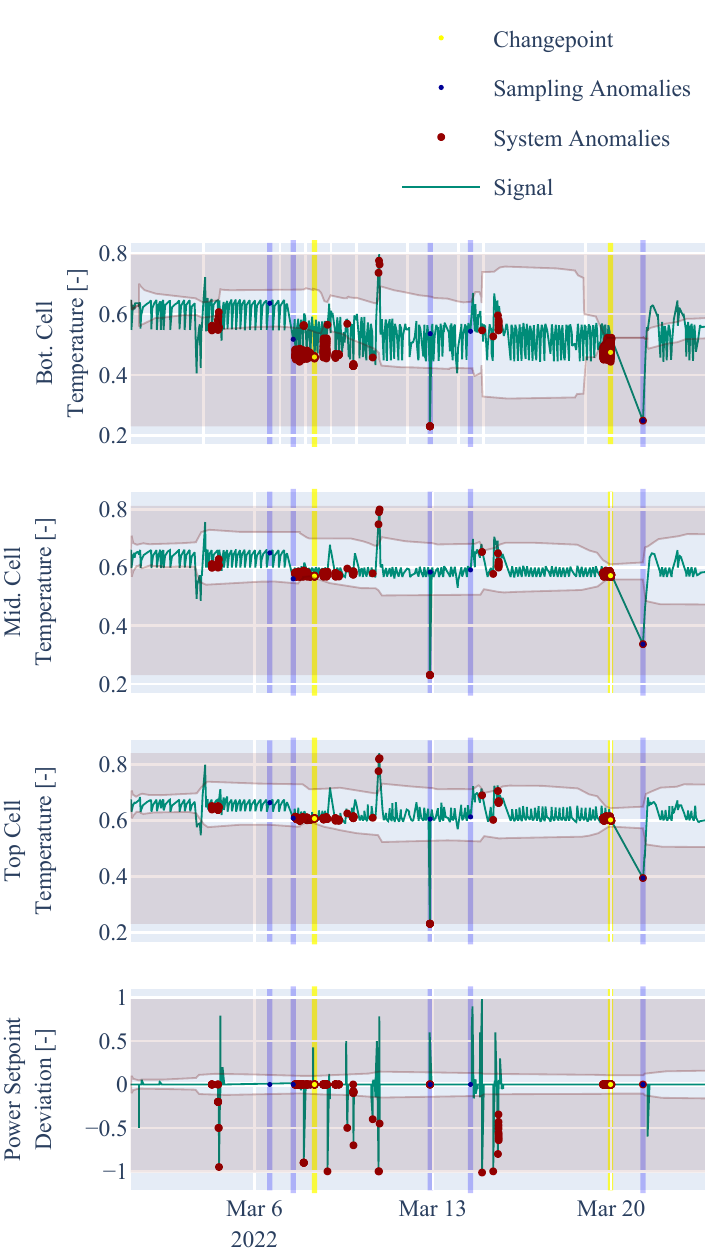}}
\caption{Time Series of BESS measurements (green line) of process variables. Non-uniform ticks on the x-axis mark days of interest (NOTE: some marks are hidden due to the readability). The y-axis renders the normalized process variables. System anomalies are marked as red dots. Non-uniform sampling is detected at blue vertical lines. Yellow vertical lines denote changepoint adaptation}
\label{fig:process}
\end{figure}

Fig. \ref{fig:process} depicts the operation of the BESS over March 2022. Multiple events of anomalous behavior were reported by the operators, that are observable through a sudden or significant shift in measurements in a given period. As the first step, the model was initialized, following Subsection \ref{init} system parameters were selected as $t_e = 4$ days and $T = -25$ for the expiration period and threshold accordingly. By design, the first 3 days are within the grace period, during which the model is calibrated.

The deployment and operation of the anomaly detection system were successful as shown by its adaptation of changepoint on 7\textsuperscript{th} March 2022 that appeared due to the relocation of the battery storage system outdoors. The model was adapted online based on Subsection \ref{train}. A dramatic shift in environmental conditions changed the dynamics of the system's temperature. However, new behavior was adopted by the top-level anomaly isolation system within two days, significantly reducing the number of alerts afterward. Interestingly, calibration of the system, as shown in deviations of setpoint from real power demand and multiple peaks in temperature were captured as well. 

The system identified 5 deviations in sampling, denoted by the blue line in Fig. \ref{fig:process}. The longest packet loss observed happened across 20\textsuperscript{th} March up to 21\textsuperscript{st}. Unexpectedly, the change point detection module identified start of the loss as a changepoint, followed by the detector of deviations in sampling that could only capture the event with the next observation after the loss. Red dots represent anomalies at the system level given by equation \eqref{eq:anomaly}. The dynamic signal limits are surpassed in one or multiple signals during the system's anomalies. Moreover, shifts in the relationship between variables are considered by the model based on the correlation matrix.

\section{Conclusion}
In this paper, we examine the capacity of joint probability distribution to model the normal operation of dynamic systems employing streaming IoT devices. The dynamics of the systems are elaborated in the model using Welford’s online algorithm with the capacity to update and revert sufficient parameters of multivariate Gaussian distribution in time making it possible to elaborate non-stationarity in the process variables.

We assume the Gaussian distribution of measurements over a bounded time frame related to the system dynamics. We consider such an assumption reasonable, with support of multiple trials where the Kolmogorov–Smirnov test did not reject this hypothesis. The statistical model provides the capacity for the interpretation of the anomalies as extremely deviating observations from the mean vector. Another assumption held in this study is that any anomaly, spatial or temporal, can be transformed in such a way that makes it an outlier given one or more interacting signals.

Our approach establishes the system's operation state at the global anomaly level by considering interactions between input measurements and engineered features. At the second level, dynamic process limits based on PPF at threshold probability, given multivariate distribution parameters, help isolate the root cause of anomalies. This level serves the diagnostic purpose of the model operation. The individual signals contribute to the global anomaly prediction, while the proposed dynamic limits offer less conservative restrictions on individual process operation. In parallel, the detector allows discrimination of signal losses due to packet drops and sensor malfunctioning.

The ability to detect and identify anomalies in the system, isolate the root cause of anomaly to specific signal or feature, and identify signal losses is shown in two case studies on real data. Unlike many anomaly detection approaches, the proposed RAID method does not require historical data or ground truth information about anomalies, relieving general limitations.

The first case study performed on benchmark industrial data showed the ability to provide comparable results to other self-learning adaptable anomaly detection methods allowing, in addition, the root cause isolation.
The second case study, performed on real operation data of BESS, examined the battery energy storage system and demonstrated the ability to capture system anomalies and provide less conservative limits to signals and extracted features.

Future works on the method will include improvement to the scalability, a decrease in the latency on high dimensional data, and false positive rate reduction, from which general plug-and-play models suffer.

\bibliographystyle{IEEEtran}
\bibliography{main}

\begin{thebibliography}{10}
\providecommand{\url}[1]{#1}
\csname url@samestyle\endcsname
\providecommand{\newblock}{\relax}
\providecommand{\bibinfo}[2]{#2}
\providecommand{\BIBentrySTDinterwordspacing}{\spaceskip=0pt\relax}
\providecommand{\BIBentryALTinterwordstretchfactor}{4}
\providecommand{\BIBentryALTinterwordspacing}{\spaceskip=\fontdimen2\font plus
\BIBentryALTinterwordstretchfactor\fontdimen3\font minus \fontdimen4\font\relax}
\providecommand{\BIBforeignlanguage}[2]{{%
\expandafter\ifx\csname l@#1\endcsname\relax
\typeout{** WARNING: IEEEtran.bst: No hyphenation pattern has been}%
\typeout{** loaded for the language `#1'. Using the pattern for}%
\typeout{** the default language instead.}%
\else
\language=\csname l@#1\endcsname
\fi
#2}}
\providecommand{\BIBdecl}{\relax}
\BIBdecl

\bibitem{Chandola2009}
\BIBentryALTinterwordspacing
V.~Chandola, A.~Banerjee, and V.~Kumar, ``Anomaly detection: A survey,'' \emph{ACM Comput. Surv.}, vol.~41, no.~3, jul 2009. [Online]. Available: \url{https://doi.org/10.1145/1541880.1541882}
\BIBentrySTDinterwordspacing

\bibitem{Laptev2015}
\BIBentryALTinterwordspacing
N.~Laptev, S.~Amizadeh, and I.~Flint, ``Generic and scalable framework for automated time-series anomaly detection,'' in \emph{Proceedings of the 21th ACM SIGKDD International Conference on Knowledge Discovery and Data Mining}, ser. KDD '15.\hskip 1em plus 0.5em minus 0.4em\relax New York, NY, USA: Association for Computing Machinery, 2015, pp. 1939--1947. [Online]. Available: \url{https://doi.org/10.1145/2783258.2788611}
\BIBentrySTDinterwordspacing

\bibitem{Kejariwal2015}
\BIBentryALTinterwordspacing
A.~Kejariwal, ``Introducing practical and robust anomaly detection in a time series,'' 2015. [Online]. Available: \url{https://blog.twitter.com/engineering/en_us/a/2015/introducing-practical-and-robust-anomaly-detection-in-a-time-series}
\BIBentrySTDinterwordspacing

\bibitem{Cook2020}
A.~A. Cook, G.~Mısırlı, and Z.~Fan, ``Anomaly detection for iot time-series data: A survey,'' \emph{IEEE Internet of Things Journal}, vol.~7, no.~7, pp. 6481--6494, 2020.

\bibitem{Fan2019}
\BIBentryALTinterwordspacing
C.~Fan, Y.~Sun, Y.~Zhao, M.~Song, and J.~Wang, ``Deep learning-based feature engineering methods for improved building energy prediction,'' \emph{Applied Energy}, vol. 240, pp. 35--45, 2019. [Online]. Available: \url{https://www.sciencedirect.com/science/article/pii/S0306261919303496}
\BIBentrySTDinterwordspacing

\bibitem{Talagala2021}
\BIBentryALTinterwordspacing
P.~D. Talagala, R.~J. Hyndman, and K.~Smith-Miles, ``Anomaly detection in high-dimensional data,'' \emph{Journal of Computational and Graphical Statistics}, vol.~30, no.~2, pp. 360--374, 2021. [Online]. Available: \url{https://doi.org/10.1080/10618600.2020.1807997}
\BIBentrySTDinterwordspacing

\bibitem{Wu2021}
H.~Wu, J.~He, M.~Tömösközi, Z.~Xiang, and F.~H. Fitzek, ``In-network processing for low-latency industrial anomaly detection in softwarized networks,'' in \emph{2021 IEEE Global Communications Conference (GLOBECOM)}, Dec 2021, pp. 01--07.

\bibitem{Ahmad2017134}
\BIBentryALTinterwordspacing
S.~Ahmad, A.~Lavin, S.~Purdy, and Z.~Agha, ``Unsupervised real-time anomaly detection for streaming data,'' \emph{Neurocomputing}, vol. 262, pp. 134--147, 2017, online Real-Time Learning Strategies for Data Streams. [Online]. Available: \url{https://www.sciencedirect.com/science/article/pii/S0925231217309864}
\BIBentrySTDinterwordspacing

\bibitem{Bosman201514}
\BIBentryALTinterwordspacing
H.~H. Bosman, G.~Iacca, A.~Tejada, H.~J. Wörtche, and A.~Liotta, ``Ensembles of incremental learners to detect anomalies in ad hoc sensor networks,'' \emph{Ad Hoc Networks}, vol.~35, pp. 14--36, 2015, special Issue on Big Data Inspired Data Sensing, Processing and Networking Technologies. [Online]. Available: \url{https://www.sciencedirect.com/science/article/pii/S1570870515001481}
\BIBentrySTDinterwordspacing

\bibitem{Carletti2019}
M.~Carletti, C.~Masiero, A.~Beghi, and G.~A. Susto, ``Explainable machine learning in industry 4.0: Evaluating feature importance in anomaly detection to enable root cause analysis,'' in \emph{2019 IEEE International Conference on Systems, Man and Cybernetics (SMC)}, Oct 2019, pp. 21--26.

\bibitem{Nguyen2019}
Q.~P. Nguyen, K.~W. Lim, D.~M. Divakaran, K.~H. Low, and M.~C. Chan, ``Gee: A gradient-based explainable variational autoencoder for network anomaly detection,'' in \emph{2019 IEEE Conference on Communications and Network Security (CNS)}, June 2019, pp. 91--99.

\bibitem{Amarasinghe2018}
K.~Amarasinghe, K.~Kenney, and M.~Manic, ``Toward explainable deep neural network based anomaly detection,'' in \emph{2018 11th International Conference on Human System Interaction (HSI)}, July 2018, pp. 311--317.

\bibitem{Yang2022}
\BIBentryALTinterwordspacing
W.-T. Yang, M.~S. Reis, V.~Borodin, M.~Juge, and A.~Roussy, ``An interpretable unsupervised bayesian network model for fault detection and diagnosis,'' \emph{Control Engineering Practice}, vol. 127, p. 105304, 2022. [Online]. Available: \url{https://www.sciencedirect.com/science/article/pii/S0967066122001502}
\BIBentrySTDinterwordspacing

\bibitem{Wadinger2023}
M.~Wadinger and M.~Kvasnica, ``Real-time outlier detection with dynamic process limits,'' in \emph{Proceedings of the 2023 24th International Conference on Process Control (PC)}, 2023, in press.

\bibitem{Wel62}
\BIBentryALTinterwordspacing
B.~P. Welford, ``Note on a method for calculating corrected sums of squares and products,'' \emph{Technometrics}, vol.~4, no.~3, pp. 419--420, 1962. [Online]. Available: \url{https://www.tandfonline.com/doi/abs/10.1080/00401706.1962.10490022}
\BIBentrySTDinterwordspacing

\bibitem{Genz2000}
A.~Genz, ``Numerical computation of multivariate normal probabilities,'' \emph{Journal of Computational and Graphical Statistics}, vol.~1, 05 2000.

\bibitem{Brent72}
\BIBentryALTinterwordspacing
R.~P. Brent, \emph{Algorithms for minimization without derivatives}.\hskip 1em plus 0.5em minus 0.4em\relax Englewood Cliffs, N.J: Prentice-Hall, 1972. [Online]. Available: \url{https://openlibrary.org/books/OL4739237M/Algorithms_for_minimization_without_derivatives}
\BIBentrySTDinterwordspacing

\bibitem{skab2020}
I.~D. Katser and V.~O. Kozitsin, ``Skoltech anomaly benchmark (skab),'' \url{https://www.kaggle.com/dsv/1693952}, 2020.

\end{thebibliography}

\end{document}